# Variational Autoencoder-Based Approach to Latent Feature Analysis on Efficient Representation of Power Load Monitoring Data

*Boyu Xie[1], Tangtang Xie[1*]*

[1]*College of Computer and Information Science, Southwest University, Chongqing, China*
* xtt@swu.edu.cn

## Abstract

With the development of smart grids, High-Dimensional and Incomplete (HDI) Power Load Monitoring (PLM) data challenges the performance of Power Load Forecasting (PLF) models. In this paper, we propose a potential characterization model VAE-LF based on Variational Autoencoder (VAE) for efficiently representing and complementing PLM missing data. VAE-LF learns a low-dimensional latent representation of the data using an Encoder-Decoder structure by splitting the HDI PLM data into vectors and feeding them sequentially into the VAE-LF model, and generates the complementary data. Experiments on the UK-DALE dataset show that VAE-LF outperforms other benchmark models in both 5% and 10% sparsity test cases, with significantly lower RMSE and MAE, and especially outperforms on low sparsity ratio data. The method provides an efficient data-completion solution for electric load management in smart grids.

## 1   Introduction

In recent years, as global environmental problems have intensified and energy resources have become increasingly limited, the imperative for energy conservation and emission reduction in power systems has become increasingly evident. The advent of smart grids has catalyzed the emergence of PLF as a pivotal research domain [1]. PLF is a critical technology for contemporary power systems. It utilizes historical power consumption data, meteorological conditions, economic activities, and other relevant factors to predict future power demand. It is the fundamental nexus of power system planning, scheduling, and operation, exerting a direct influence on the stability, economy, and reliability of the power grid.

PLM pertains to the real-time or periodic collection and analysis of electricity consumption data in the power system [2], [3]. This is done to ensure the safe, stable and efficient operation of the power system, by monitoring load changes, operation status and abnormalities. This technology constitutes a pivotal component within the realm of smart grids and energy management, finding extensive application in domains such as power scheduling, equipment maintenance, and the optimisation of energy efficiency. It provides a broad data base for PLF models. However, due to the presence of sampling sparsity, sensor limitations, monitoring equipment failures, transmission channel congestion and anomalies, which lead to missing power load monitoring data, the PLM data is therefore HDI data [4]-[20]. This has the potential to directly affect the performance of PLF models. Accurate data representation and missing value management are critical research issues in this field.

Among them, Matrix Factorization [21]-[27] (MF)-based Latent Feature Analysis (LFA) [28]-[67] models have attracted much attention for their success in deriving accurate latent features from HDI data. These models transform the features of each node into a low-dimensional latent space through MF and estimate missing data by computing the inner product between nodes. However, such models are linear, which leads to limitations in their representation capabilities, on the other hand, Neural Network (NN)-based methods excel at extracting intricate nonlinear features and have been increasingly adopted for LFA tasks with HDI data. Examples include Autoencoder (AE)-based models [68]-[71], Neural MF-based models [72], Variational Graph Autoencoder-based models [73], Graph Neural Network (GNN)-based models [74] and Graph Convolutional Network (GCN)-based models [75]-[80].

AE learns low-dimensional representations of data through an encoder-decoder structure, but its hidden variables lack explicit probabilistic meaning, limiting generative power. VAE is a generative model, it is proposed by Kingma et al. [81] in 2014. VAE combines NN and Bayesian inference for learning latent representations of data and generating new samples. We use VAE to complement PLM missing data by first splitting the PLM data into vectors, and then inputting the vectors sequentially to VAE for imputation. The primary contributions of this study are outlined below:

1) Aiming at the high-dimensional incompleteness problem of PLM data, we propose the VAE-LF model, which introduces the VAE to the task of interpolating PLM missing data. By combining neural networks and Bayesian inference, VAE-LF is able to effectively learn the nonlinear latent features of the data.

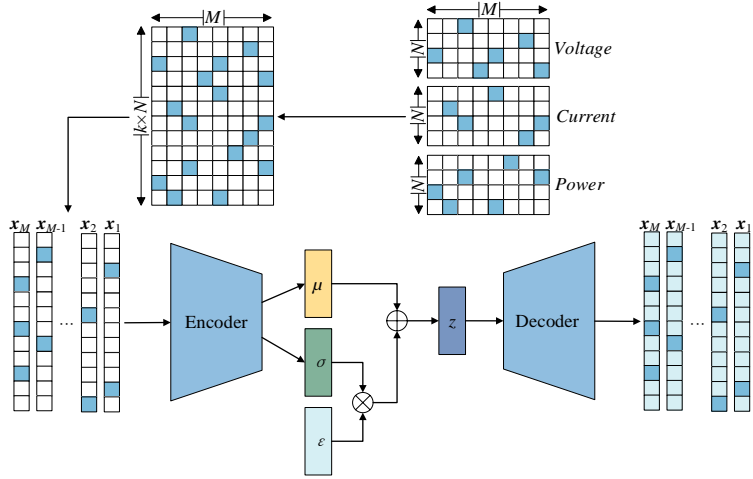

Fig. 1 An illustration of VAE-LF's overall structure.

2) We design a novel approach to PLM data processing by splitting the high-dimensional time-days matrix into vectors, which are sequentially input into the VAELFA model for imputation. This approach makes full use of the serialization processing capability of VAE and adapts to the temporal characteristics and sparsity of PLM data.

## 2. Preliminaries

*Definition* **1** *(HDI PLM data)*: PLM data involves data that includes measured parameters such as voltage, current, power, etc., sampled at fixed moments each day. For each parameter, it is sampled M times per day for a total of N days, which results in a time-days matrix of dimension |N|×|M|. Assuming that a total of k parameters are sampled, this results in k time-days matrices, which are spliced along the time dimension, resulting in a Xk|N|×|M| HDI matrix.

For known and unknown data, denoted by Λ and Γ respectively. If |Λ|<<|Γ|, then X is an HDI data.

*Definition* **2** *(LFA Model)*: For an input HDI matrix X, the LFA model produces a rank $D$ approximation $\hat{X}$, where each $\hat{x}_{n,m}$ corresponds to $x_{n,m} \in X$, and the latent dimension D is much smaller than min{|N|, |M|}.

To acquire the targeted latent features and estimations, the objective function grounded in Euclidean distance, which is generally employed, is defined solely on to measure the difference between X and $\hat{X}$, as shown in equation:

$$\varepsilon = \frac{1}{2} \sum_{x_{n,m} \in \Lambda} \left( x_{n,m} - \hat{x}_{n,m} \right)^2 \quad (1)$$

## 3. Methodology

The structure of VAE-LF is illustrated in Figure 1. VAE-LF comprises two components: an Encoder and a Decoder. The Encoder transforms the input data x into a probability distribution, typically a Gaussian distribution, of the latent variable z. The Decoder is used to generate a new sample, which generates an approximation of the x by reconstructing it from the z. The k-parameter time-days matrix of dimension |N|×|M| is first spliced into matrix $X^{k|M|\times|N|}$ along the time dimension. The matrix X is then split into *M* vectors along the time dimension, and these vectors are sequentially input into the VAE-LF for missing data imputation.

*3.1 Variational Autoencoder*

The VAE framework comprises an inference network (Encoder) and a generative network (Decoder): The Encoder approximates the posterior $p_\theta(z|x)$ as $q_\phi(z|x)$. The Decoder maximizes the likelihood $p_\theta(x|z)$ for reconstruction.

*3.1.1 Basic Probability Model:* suppose we have a dataset X = {$x_1, x_2, \cdots, x_N$}, and we wish to model the process of generating the data by means of the latent variable *z*. The joint probability of generating the model is:

$$p_\theta(x,z) = p_\theta(x|z)p_\theta(z) \quad (2)$$

where $p_\theta(z)$ denotes the prior distribution of the latent variable *z*, generally presumed to conform to a standard normal distribution $\mathcal{N}(0, I)$, and $p_\theta(x|z)$ is the conditional probability, indicating the likelihood of producing the *x* given the *z*. Our goal is to maximize the marginal likelihood $p_\theta(x)$ of the data:

$$p_\theta(x) = \int p_\theta(x|z) p_\theta(z) dz \quad (3)$$

However, it is usually not feasible to compute this integral directly because of the high dimensionality and complex distribution of the potential space. Therefore, the VAE employs variational inference to estimate the posterior distribution $p_\theta(z|x)$.

*3.1.2 Approximate posteriori:* to estimate the posterior $p_\theta(z|x)$, we introduce a variational distribution $q_\phi(z|x)$, and we generally use neural networks to parameterize the approximate posterior. For example, assuming that the posterior distribution satisfies a Gaussian distribution with homogeneous terms, we have:

$$q_\phi(z|x) = \mathcal{N}(z; \mu(x), \sigma^2(x)I) \tag{4}$$

where $\mu(x)$ and $\sigma(x)$ represent the mean and standard deviation produced by the encoder.

*3.1.3 Evidence Lower Bound (ELBO):* the goal of VAE is to optimize the model by minimizing the KL divergence between $q_\phi(z|x)$ and $p_\theta(z|x)$, the $p_\theta(z|x)$ represent the true posterior, and the objective function is obtained by maximizing the marginal likelihood of the observed variable $x$, which is derived as follows:

$$\begin{aligned}
\log p_\theta(x) &= \log p_\theta(x,z) - \log p_\theta(z|x) \\
&= \log \frac{p_\theta(x,z)}{q_\phi(z|x)} + \log \frac{q_\phi(z|x)}{p_\theta(z|x)} \\
&= \underbrace{\mathbb{E}_{q_\phi(z|x)}\left[\log \frac{p_\theta(x,z)}{q_\phi(z|x)}\right]}_{\mathcal{L}_{\theta,\phi}(x)} + \underbrace{\mathbb{E}_{q_\phi(z|x)}\left[\log \frac{q_\phi(z|x)}{p_\theta(z|x)}\right]}_{D_{KL}(q_\phi(z|x) \| p_\theta(z|x)) \geq 0} \\
&\geq \mathbb{E}_{q_\phi(z|x)}\left[\log \frac{p_\theta(x,z)}{q_\phi(z|x)}\right]
\end{aligned} \tag{5}$$

where the second term represents the Kullback-Leibler divergence between $q_\phi(z|x)$ and $p_\theta(z|x)$, while the first term constitutes the lower bound of $\log p\theta(z|x)$, known as the ELBO. Therefore the ELBO is:

$$\text{ELBO} = \mathbb{E}_{q_\phi(z|x)}\left[\log p_\theta(x|z)\right] - D_{KL}\left(q_\phi(z|x) \| p_\theta(z|x)\right) \tag{6}$$

Since the KL divergence is non-negative, optimizing the ELBO corresponds to maximizing the lower bound of $\log p_\theta(z|x)$.

*3.1.4 Loss function:* the first term in Equation (6) represents the reconstruction loss, which represents the expected likelihood of reconstructing x from z. It is usually assumed that $p_\theta(z|x)$ is Gaussian distributed, and the logarithm of the likelihood can be simplified to the mean square error. The second KL divergence measures the difference between the variational distribution and the prior distribution and acts as a regularizer. Assuming a prior $p_\theta(z) = \mathcal{N}(0, I)$ and a variational distribution $q_\phi(z|x) = \mathcal{N}((\mu(x), \sigma(x)I)$, the analytic form of the KL divergence is:

$$D_{KL}\left(\mathcal{N}(\mu, \sigma^2) \| \mathcal{N}(0,1)\right) = \frac{1}{2}\sum_{i=1}^{d}\left(\mu_i^2 + \sigma_i^2 - \log \sigma_i^2 - 1\right) \tag{7}$$

where $d$ denotes the dimensionality of the latent variable. The reconstruction loss is usually approximated as:

$$\mathbb{E}_{q_\phi(z|x)}\left[\log p_\theta(x|z)\right] \approx -\frac{1}{2}\|x - \hat{x}\|^2 \tag{8}$$

where $\hat{x}$ represents the data reconstructed from z by the decoder. Thus, the VAE's loss function is formulated as follows:

$$\mathcal{L} = \frac{1}{2}\|x - \hat{x}\|^2 + \frac{1}{2}\sum_{i=1}^{d}\left(\mu_i^2 + \sigma_i^2 - \log \sigma_i^2 - 1\right) \tag{9}$$

*3.1.5 Reparameterization Trick:* Kingma et al. [81] cleverly used the Reparameterization Trick to make ELBO straightforward to derive. We use the Reparameterization Trick to sample z.

$$z = \mu(x) + \sigma(x) \cdot \varepsilon, \quad \varepsilon \sim \mathcal{N}(0, I) \tag{10}$$

In this way, the sampling process of z is converted into a deterministic computation, allowing backpropagation to refine the parameters of the encoder and decoder.

*3.2 Encoder*

The encoder receives one input vector $x_m$ at a time, and the input vector is first encoded into a hidden variable representation h through a fully connected layer, followed by another fully connected layer to parameterize the mean $\mu$ and standard variance $\sigma$ of a Gaussian distribution. The hidden variable z can then be sampled from a Gaussian distribution $\mathcal{N}(\mu, \sigma)$ by applying the reparameterization technique, and the whole encoder can be expressed as follows:

$$h = a_1(w_1 x_m + b_1)$$
$$\mu = a_2(w_2 h + b_2)$$
$$\sigma = a_3(w_3 h + b_3) \qquad (11)$$
$$z = \mu + \sigma \otimes \varepsilon$$

where $a_1(\cdot)$, $a_2(\cdot)$ and $a_3(\cdot)$ represent the activation function of the first, second and third layers respectively, we use the ReLU function here. The $\varepsilon$ is sampled from $\mathcal{N}(0, I)$.

*3.3 Decoder*

The decoder utilizes a fully-connected layer to reconstruct the input data $x$ from the latent variable $z$ to generate an approximation. The decoder output is defined as:

$$\hat{x} = g(w_5 a_4(w_4 z + b_4) + b_5) \qquad (12)$$

where $a_4(\cdot)$ denotes the ReLU function, and the $g(\cdot)$ represent the activation function of the last layer, here we use the sigmoid activation function.

# 4 Experimentation

*4.1 General Settings*

*4.1.1 Datasets:* we used a widely used PLM dataset UK-DALE in our experiments. and we divided the dataset into test cases with sparsity ratio of 5% and 10%. For example, the test cases with a sparsity ratio of 5% show that only 5% of the entries in the target tensor are known and the remaining 95% are unknown. These datasets record load monitoring data from multiple appliances in the user's home. Considering the cyclical nature of residential electricity consumption, we selected 21 consecutive days of data from these two datasets and collected 86,400 samples per day for monitoring parameters including power, voltage and apparent power. We divided the datasets into non-overlapping training (60%), validation (20%), and test (20%) subsets.

*4.1.2 Evaluation metrics:* the purpose of this study is to assess the representation learning capability of test models about PLM data. Typically, this ability is measured by the model's ability to predict missing information in PLM data. RMSE and MAE were selected as the principal metrics to assess model performance. These metrics are widely used to measure the accuracy and error level of the model in prediction tasks and provide an intuitive assessment of the model performance. It should be noted that $x_{n,m}$ describes the actual parameters such as voltage, power, etc. used in the real world, and the lower the value of RMSE, the more accurate the predicted data.

$$RMSE = \sqrt{\sum_{x_{n,m} \in \Omega} (x_{n,m} - \hat{x}_{n,m})^2 / |\Omega|}$$
$$MAE = \sum_{x_{n,m} \in \Omega} |x_{n,m} - \hat{x}_{n,m}| / |\Omega| \qquad (13)$$

*4.1.3 Comparison Models:* to evaluate the effectiveness of VAE-LF in predicting missing PLM data, we compare VAE-LF with three representative missing data imputation methods. Their details are as follows:

*M1. VAE-LF*: the model proposed in this paper.

*M2. HMLET* [82]: improves the accuracy of GNN-based recommender systems by dynamically selecting linear or nonlinear propagation steps, combined with gating mechanisms and residual prediction.

*M3. GTN* [83]: by introducing graph trend filtering techniques to capture the adaptive reliability of user-item interactions, the robustness and performance of recommender systems are significantly improved.

*M4. LightGCN* [80]: significantly improves collaborative filtering performance in recommender systems by simplifying graph convolutional networks and retaining only neighborhood aggregation and layer combination.

*4.2 Results*

In order to evaluate the effectiveness of the model in performing PLM missing data imputation, here we evaluate the proposed model against three notable models for predicting missing data. Table 1 presents the RMSE and MAE outcomes for models M1–M4 evaluated on datasets D1–D2, where the sparsity ratio is 5% for the D1 and 10% for the D2. The Figure 2 depicts the results of Table 1. From the table it can be seen that on D1 the RMSE of M1 reaches a minimum of 0.1384. which is about 8.16% lower than the 0.1507 of M2, about 20.18% lower than the 0.1734 of M3, and about 24.58% lower than the 0.1835 of M4. On the RMSE of D2, M1 improves by 3.51%, 17.65% and 7.47% compared to M2-M4, respectively. Similar results are observed in other test cases.

In addition, D1 has a lower sparsity ratio than D2. It can be observed that the RMSE and MAE values for M1 on D1 are lower than those on D2 respectively and the improvement on D1 is more significant compared to other models. Therefore, we conclude that our proposed model outperforms other benchmark models in accurately estimating missing values of HDI PLM data and is more advantageous on low-ratio datasets.

Table 1 RMSE and MAE of M1-M4 on D1-D2.

| Dataset | Metric | M1 | M2 | M3 | M4 |
|---|---|---|---|---|---|
| D1 | RMSE | **0.1384** | 0.1507 | 0.1734 | 0.1835 |
|    | MAE  | **0.0820** | 0.1072 | 0.1268 | 0.1664 |
| D2 | RMSE | **0.1400** | 0.1451 | 0.1700 | 0.1513 |
|    | MAE  | **0.0822** | 0.1033 | 0.1255 | 0.1093 |

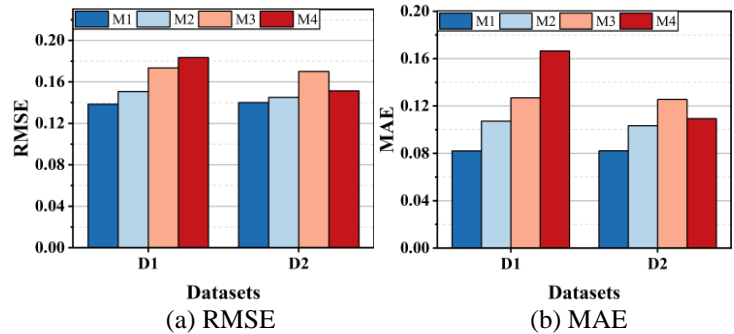

(a) RMSE  (b) MAE

Fig. 2 The comparison results of M1-M4 on D1-D2.

## 5 Conclusion

We propose the VAE-LF model to achieve efficient missing data imputation for the high-dimensional incompleteness problem in PLM data. By partitioning the PLM data into vectors and inputting them into the VAE-LF model, we successfully extracted the nonlinear latent features of the data and generated high-quality complementary data by utilizing its encoder-decoder structure and variational inference capability. Experimental validation on UK-DALE dataset for test cases with different sparsity ratio (5% and 10%) underscores that our model surpasses other comparative models in both RMSE and MAE, and especially exhibits stronger robustness and accuracy on sparsity ratio datasets. These findings highlight the efficacy of VAE-LF in processing HDI PLM data to support power scheduling and energy management in smart grids. Future work will explore more complex variants of VAE, optimize the model's performance on higher sparsity and diverse datasets, and incorporate real-time data streaming to further enhance its utility.